\documentclass[12pt]{article}

\usepackage{amsmath, amsfonts, amssymb, bm, amsthm}
\usepackage{graphicx,psfrag} %
\usepackage{enumerate}
\usepackage{natbib}
\usepackage{xcolor}
\usepackage{url} %
\usepackage[letterpaper,margin=1in]{geometry}
\usepackage{setspace}
\usepackage[hidelinks]{hyperref}

\newcommand{\blind}{1}

\graphicspath{{images/}}

\newtheorem{theorem}{Theorem}
\newtheorem{proposition}{Proposition}
\newtheorem{remark}{Remark}
\newtheorem{corollary}{Corollary}
\newtheorem{assumption}{Assumption}

\newcommand{\tr}{\mathrm{tr}}

\begin{document}

\if1\blind
{
  \title{\bf Prediction-Powered LLM Ranking with Adaptively Collected Human Feedback}
 \author{
 Gabriel Sargent\thanks{Department of Statistics and Operations Research, University of North Carolina at Chapel Hill. Email: gsargent@unc.edu.}\qquad
 Will Wei Sun\thanks{Daniels School of Business, Purdue University. Email: sun244@purdue.edu.}\qquad
 Zhengwu Zhang\thanks{Department of Statistics and Operations Research, University of North Carolina at Chapel Hill. Email: zhengwu\_zhang@unc.edu.}\qquad
 Yufeng Liu\thanks{Department of Statistics, University of Michigan. Email: yufliu@umich.edu.}\qquad
 }
 \date{}
  \maketitle
} \fi

\if0\blind
{
  \bigskip
  \bigskip
  \bigskip
  \begin{center}
    {\LARGE\bf Prediction-Powered LLM Ranking with Adaptively Collected Human Feedback}
  \end{center}
  \medskip
} \fi

\begin{singlespace}
\begin{abstract}

Ranking large language models (LLMs) through human evaluations is essential for downstream decision-making. Because human feedback is expensive, evaluation platforms often collect it \emph{adaptively}, prioritizing informative model comparisons using previously observed feedback. To further improve precision with limited human feedback, evaluators can draw on abundant unrated responses and relatively inexpensive LLM judges; however, judge bias and adaptive dependence challenge standard i.i.d. inference frameworks. We propose Prediction-Powered Ranking (PP-Rank), a framework that combines LLM-judge predictions with human feedback to improve precision while providing valid inference under adaptive sampling. Data-driven prediction weights minimize the asymptotic covariance trace within the proposed class. In theory,  we establish asymptotic normality for general evaluation targets and simultaneous ranking inference. A joint martingale analysis handles dependence between adaptive human feedback and auxiliary data. We further derive a rule for reducing exploration that prioritizes lower-cost comparisons while matching a conservative variance bound, and characterize allocation limits for regular $\sqrt{T}$-rate estimation of pairwise means. Semi-synthetic experiments on UltraFeedback-Extended show improvements over human-only baselines in score estimation, confidence interval width, and ranking certification. Synthetic experiments demonstrate robustness to judge quality and adaptive allocation, while showing that PP-Rank can substantially reduce evaluation costs without sacrificing precision.

\end{abstract}

\noindent


\end{singlespace}

\vfill
\newpage

\doublespacing

\baselineskip=25pt %
\section{Introduction}
\label{sec:intro}
Reliably ranking LLMs through human evaluations \citep{Arena} is essential for downstream decision-making: rankings inform which models users and organizations deploy, which models developers  invest in, and policy responses to the evolving capabilities of  particular classes of models, such as open-weight LLMs.
 A central bottleneck in human evaluation, however, is that reliable inference typically requires many human preferences, yet collecting such feedback is slow and expensive.
Consequently, limited human feedback may leave substantial uncertainty about relative model performance, leaving decision-makers poorly equipped to make these choices. 

Due to this scarcity, human feedback in LLM evaluation is often collected \emph{adaptively}; evaluators use evolving information to allocate human feedback where it is most valuable \citep{Arena,  InclusionArena, CostOptimal, HUMAINE, gu2026anytimevalid, li2026low}.
For example, adaptive sampling can concentrate comparisons among closely ranked models to resolve uncertainty, oversample newly introduced models to assess them rapidly, and favor highly ranked models to maintain user satisfaction.
Yet even with adaptive allocation, limited human feedback may leave substantial uncertainty about model rankings. An October 2026 snapshot of Arena's math leaderboard illustrates this uncertainty: the highest-ranked model had a rank set spanning 1 to 29, and the rank sets for the top five models contained 27 possible ranks on average \citep{ArenaMathLeaderboard}.

A complementary way to improve precision is to leverage auxiliary information.
LLM evaluation settings often provide two such sources: model responses without human ratings and relatively inexpensive feedback from LLM judges.
Specifically, evaluators also have access to large collections of responses across many tasks, even when only a small fraction have been scored by humans, while strong LLM judges can provide predicted human feedback for the unrated examples.
However, integrating these sources poses two key statistical challenges. First, LLM-judge ratings can be systematically misaligned with human preferences, which introduces bias.
Second, adaptive model-pair selection violates the i.i.d. assumptions underlying standard prediction-powered methods. Methods that ignore unequal comparison probabilities give greater influence to frequently sampled model pairs. When the working preference model is misspecified, this can lead to estimated rankings that differ from the intended population ranking, even with arbitrarily large samples.

\begin{figure}
\begin{center}
\includegraphics[width=6.25in]{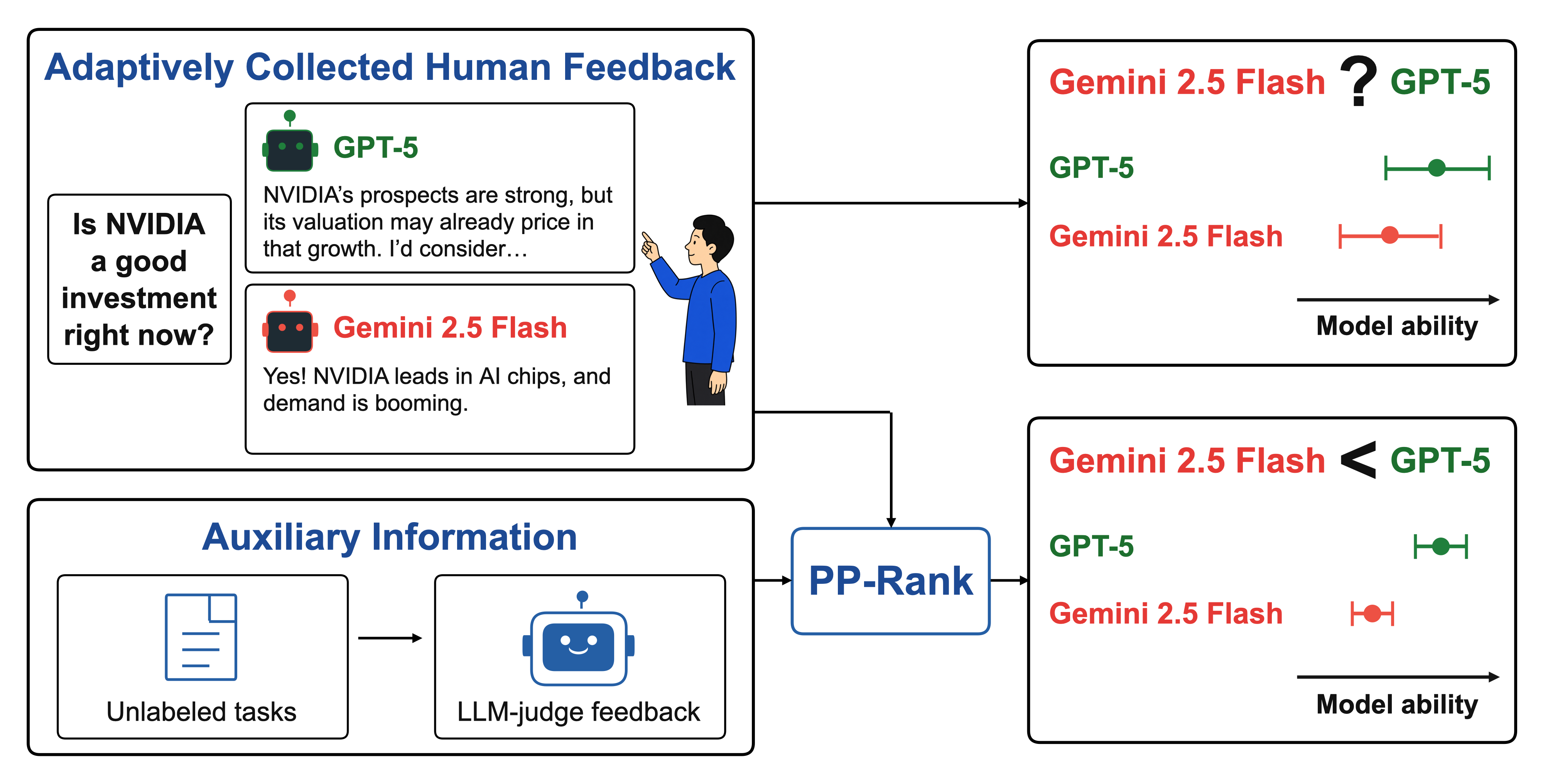}
\end{center}
\caption{Human feedback is often too scarce to precisely compare models, leaving decision-makers unable to certify conclusions about their relative abilities.
PP-Rank additionally leverages unlabeled tasks and LLM-judge feedback to reduce uncertainty, enabling such conclusions to be certified. (Model responses shown are hypothetical.)
}\label{fig: PP-Rank}
\end{figure}

To address these challenges, we introduce Prediction-Powered Ranking (PP-Rank), an inferential framework that integrates LLM-judge ratings, unlabeled model responses, and adaptively collected human feedback (see Figure \ref{fig: PP-Rank}).
PP-Rank therefore enables evaluators to leverage auxiliary information for valid and efficient inference under the adaptive sampling schemes increasingly used in practice.
We consider an LLM-ranking setting with \(K\) models. At each round \(t\), an evaluation platform observes a prompt \(\bm X_t\), selects a model pair \(A_t\) according to an adaptive pair-selection policy, and receives human feedback \(Y_t\).
The platform also has access to a large pool of unlabeled prompts and an LLM judge that predicts human preferences.
Our goal is to rank the models and quantify uncertainty in their relative performance. To this end, we construct the PP-Rank estimator \(\hat{\bm\theta}_T\) of an evaluation target \(\bm\theta^*\), such as a vector of Bradley--Terry scores, and use inference on \(\bm\theta^*\) to draw conclusions about the model ranks. Theoretically, we establish asymptotic normality of our proposed estimator under mild conditions on the adaptive data policy (Theorem \ref{thm: asymptotic}). The proof applies a martingale central limit theorem to a prediction-powered estimating equation formed from adaptively sampled observations. This result enables asymptotically valid confidence intervals and hypothesis tests for a broad class of $Z$-functionals of the feedback distribution. 

Our estimator also automatically adjusts to the predictive quality of the LLM judge by weighting the judge's predictions using a tuning parameter $\bm\lambda \in \mathbb R^{\binom K2}$. 
We characterize the variance-minimizing vector of weights $\bm\lambda^*$ and construct a consistent data-driven estimator $\hat{\bm\lambda}_T.$
This weighting enables our estimator to achieve the minimum asymptotic variance within a class of prediction-powered estimators (Corollary \ref{cor: minimalvar}).
For comparison, we consider an analogous estimator that uses only labeled observations, which we refer to as the labeled-only baseline.
When the LLM judge is informative, our estimator exploits the unlabeled prompts and LLM-judge feedback to reduce asymptotic variance.
When the predictions are uninformative, the weighting procedure shrinks their contribution, and our estimator recovers the labeled-only estimator.
Consequently, our estimator's reliance on LLM judge scores is asymptotically safe: its asymptotic variance is never larger than that of the labeled-only estimator, and our estimator achieves substantial efficiency gains when they are informative.
  
We also study how LLM-judge feedback and unlabeled prompts affect the sampling requirements for reliable rankings. We show that the better the LLM judge approximates human feedback, the less random sampling is needed to reach a given level of certainty about the leaderboard. We develop an online procedure that learns the required amount of random sampling from accumulating data and adjusts the sampling policy accordingly. Moreover, in the idealized case of a perfect LLM judge, PP-Rank remains valid even as random sampling vanishes. Conversely, when the judge is imperfect, we show that continued random sampling is necessary for any regular estimator. These results allow evaluation platforms to favor lower-cost or otherwise preferable model comparisons without compromising the reliability of their ranking conclusions.

In summary, we make the following contributions:

\begin{enumerate}

    \item \textbf{Method.}
    We develop a prediction-powered framework for LLM evaluation with adaptively collected human feedback.
    Unlike standard prediction-powered methods, PP-Rank remains valid when the task or model pair selected depends on previously observed feedback.
    PP-Rank accounts for both adaptive sampling and LLM-judge bias, enabling valid and efficient ranking.
    We develop a data-driven procedure for weighting LLM-judge feedback that makes PP-Rank robust to two important sources of variance in online LLM evaluation: inaccurate judge feedback and infrequently sampled comparisons.
    The resulting estimator has minimal asymptotic variance within our prediction-powered class; in particular, its use of auxiliary information never increases asymptotic variance and can substantially reduce it when the judge is informative.
    We also develop an online procedure that sequentially learns the amount of exploration required for reliable inference, allowing platforms to reduce spending on costly model runs and limit comparisons that degrade user satisfaction.

    \item \textbf{Theory.}
    We establish asymptotic normality for a broad class of LLM evaluation targets under adaptive data collection, including win rates, Bradley--Terry scores, and mean ratings.
    Because the labeled observations are adaptively collected, standard prediction-powered arguments based on i.i.d.\ data do not apply; our analysis instead uses martingale limit theory to establish asymptotic normality and attain valid confidence intervals and hypothesis tests.
    We also characterize how the quality of LLM-judge feedback determines the amount of random sampling required for valid leaderboard inference.
    Moreover, we prove that under realistic conditions, this sampling rate cannot be reduced to zero without compromising valid inference, regardless of LLM-judge quality, establishing a fundamental limit on the extent to which LLM-judge feedback can replace human evaluations.

    \item \textbf{Application.}
    In semi-synthetic experiments based on real prompts, LLM responses, and LLM-judge scores, we demonstrate that PP-Rank maintains valid inference under adaptive sampling, and substantially improves precision relative to using human feedback alone.
    These improvements persist across several adaptive sampling policies and enable more frequent certification of ranking statements, including pairwise model orderings and top-$K$ membership.
    We also demonstrate that PP-Rank can substantially reduce evaluation costs while maintaining inferential precision, achieving cost reductions of up to 53\% in our synthetic experiments.

\end{enumerate}

\section{Related Work}

Our work connects three related literatures: statistical methods for LLM evaluation and ranking, prediction-powered inference, and inference with adaptively collected data.

\subsection{LLM Evaluation and Ranking}
Recent statistical work develops uncertainty quantification for LLM evaluation and ranking.
\citet{DaiPromptRanking2026} construct confidence sets for prompt-dependent rankings, while \citet{li2026low} exploit low-rank structure to share information across tasks and conduct task-specific ranking inference. For sequential evaluation, \citet{gu2026anytimevalid} develop anytime-valid ranking inference under adaptive comparisons. These approaches base ranking inference solely on human pairwise comparisons.
\citet{xu2026judge} instead model heterogeneity among LLM judges, jointly estimating model quality and judge reliability without human reference labels.
Our PP-Rank focuses on improving inference for a global human-preference ranking by combining adaptively collected human comparisons with potentially biased LLM-judge predictions.

\subsection{Prediction-Powered Inference}

Prediction-powered inference (PPI) combines a small labeled sample with predictions on a large unlabeled sample to improve statistical precision \citep{PPI}.
PPI++ introduces a power-tuning procedure that adjusts the contribution of predictions to improve efficiency \citep{PPI++}.
A growing literature applies these ideas to AI evaluation, combining inexpensive judge predictions with human feedback to obtain valid and more precise estimates \citep{PPRanking, StratifiedPPI, AutoEvalDoneRight,  CostOptimal, NoisyLLMJudge, MultiplePPI, CollabEval, zhang2026conditionalevaluationlanguagemodels}. Most of these works assume an i.i.d. or otherwise fixed sampling scheme, under which tasks or model pairs are selected independently of previously observed preferences.
Recent PPI extensions accommodate adaptive label acquisition and contextual bandit inference \citep{ASI, CostOptimal, PPAI}.
These adaptive approaches primarily study selective label acquisition or inference for action-specific targets.
Our framework addresses global LLM ranking under adaptive model-pair selection, where the population target is jointly defined by pairwise comparisons across model pairs and inference remains valid even when the working preference model is misspecified.

\subsection{Inference with Adaptively Collected Data}

A complementary literature develops valid inference when data collection adapts to previously observed outcomes \citep{Deshpande2018,   ZhangBatched, Hadad2021, khamaru2024inferenceupperconfidencebound, Smith2024}.
For contextual bandits, \citet{RuiSong} study inference for linear working reward models, \citet{chang2026online} develop online inference through weighted stochastic gradient descent, and \citet{zhang2021statistical} establish an M-estimation framework for adaptively collected data.
Related work addresses model misspecification \citep{guo2025}, while \citet{shi2025statistical} reviews statistical inference in reinforcement learning.
Our setting additionally requires a joint analysis of adaptive human feedback and an auxiliary sample available to the comparison policy.
When the policy uses this sample, the human-correction and auxiliary-prediction terms cannot be treated as independent.
We establish joint asymptotic normality and consistent covariance estimation for the coupled ranking scores, while showing that consistent estimation of the prediction weights from the same observations is first-order negligible.
This analysis supports data-driven prediction tuning and uncertainty quantification using both sources of information.

\section{Problem Setting} 

We study an Arena-style online LLM-ranking experiment \citep{Arena} with $K$ models indexed by $[K]:=\{1,\dots,K\}$ and model-pair set
$\mathcal{A}:=\{(i,j):1\leq i<j\leq K\}.$ 
At each round $T,$ the data are generated as follows:
\begin{enumerate}
    \item A user comes to the evaluation platform and submits a prompt context $\bm X_t\in\mathcal X$ drawn from a fixed distribution $\mathcal P_X$, independently from historical prompts. The platform also has access to an auxiliary sample of $N$ unlabeled prompts,
$\{\tilde{\bm X}_1,\dots,\tilde{\bm X}_N\},$ drawn i.i.d. from $\mathcal{P}_X.$
 
    \item The evaluation platform chooses a pair of models
    $A_T=(I_T,J_T)\in\mathcal{A}$ according to an adaptive pair-selection policy
    $\pi_T(\cdot|\bm X_T,\mathcal{F}_{T-1}) \in \Delta(\mathcal{A}),$
    where $\mathcal{F}_{T-1}=\sigma(
\{\bm X_t,A_t,Y_t\}_{t=1}^{T-1},\allowbreak
\{\tilde{\bm X}_i\}_{i=1}^{N(T-1)})$ denotes the information accumulated before time $T$, and $\Delta(\mathcal{A})$ is the set of probability distributions over the model pairs in $\mathcal{A}.$ 
    
    \item The platform receives human feedback $Y_T.$
    For each $(i,j)\in\mathcal{A},$ let $Y_T(i,j)$ denote the potential human preference that would have been observed had models $i$ and $j$ been selected. For binary preferences, $Y_T(i,j)\in\{0,1\}$, where $Y_T(i,j)=1$ indicates that model $i$ is preferred to model $j.$ The observed preference is therefore
    $Y_T=Y_T(I_T,J_T).$
\end{enumerate}

Alongside the adaptively collected human feedback, the evaluation platform has access to an externally trained LLM judge
$\bm f=(f_{ij}(\cdot))_{(i,j)\in\mathcal{A}},$
where $f_{ij}(\bm x)$ predicts the human preference between models $i$ and $j$ under prompt $\bm x.$
Human preferences define our target of inference, while judge
predictions and auxiliary prompts provide additional information
for estimating it.

We first describe the Bradley-Terry score target, which serves
as our main example, and then introduce a general formulation of evaluation targets. For the Bradley--Terry target, let
$\bm\beta=(\beta_1,\ldots,\beta_K)^\top$ denote the model scores.
The working preference probability for pair $(j,k)$ is
$\sigma(\beta_j-\beta_k)$, where $\sigma(u)=(1+e^{-u})^{-1}$.
Because these probabilities depend only on score differences,
a common additive shift in all scores leaves them unchanged.
To remove this indeterminacy, we impose the normalization
$\sum_{i=1}^K\beta_i=0$ and use the free parameter
$\bm\theta=(\beta_1,\ldots,\beta_{K-1})^\top\in\mathbb R^{K-1}$.
The full score vector is therefore
\(
\bm\beta(\bm\theta)
=
\left(
\theta_1,\ldots,\theta_{K-1},
-\sum_{i=1}^{K-1}\theta_i
\right)^\top.
\)
The corresponding estimating function is
\begin{equation}
\bm g(\bm X,(j,k),Y(j,k);\bm\theta)
=
\big\{Y(j,k)-\sigma(\beta_j-\beta_k)\big\}
\nabla_{\bm\theta}(\beta_j-\beta_k),
\label{eqn:BTL_g}
\end{equation}
where $\bm\beta=\bm\beta(\bm\theta)$. The residual in \eqref{eqn:BTL_g} measures the discrepancy between
the observed human preference and the working preference
probability, while the gradient expresses its contribution
in terms of the free score parameters.
At the population target, the expected score contributions
balance across model pairs.

This example motivates a general formulation based on population
estimating equations.
Specifically, we conduct inference on an evaluation target $\bm\theta^\star\in\Theta\subseteq\mathbb R^d$ characterized by
\begin{equation}
\sum_{a \in \mathcal{A}}\mathbb E\big[\bm g(\bm X, a,Y(a); \bm\theta^\star)\big]=\bm 0,
\label{eqn:Z-estimation}
\end{equation}
where $\bm g$ is a known $d$-dimensional estimating function.
The expectation is taken over the fixed population of prompt
contexts and potential human preferences, and the sum assigns
equal weight to each model pair. Taking $\bm g$ as in \eqref{eqn:BTL_g} recovers the
Bradley-Terry target with $d=K-1$, whose ranking is determined
by the full score vector $\bm\beta^\star=\bm\beta(\bm\theta^\star)$. For another example, the win rate of a particular pair $a_0$
is obtained by setting
\(
g(\bm X,a,Y(a);\theta)
=
\mathbf 1_{\{a=a_0\}}\{Y(a)-\theta\},
\)
which gives $\theta^\star=\mathbb E[Y(a_0)]$.
The same framework also accommodates regression parameters
describing how preferences vary across tasks, as well as
average rubric scores when rubric ratings are observed.

This population formulation accommodates misspecification of the
working outcome model.
Under the sampling and regularity conditions stated below, our
inference remains valid for the target in
\eqref{eqn:Z-estimation}, even when the working model does not
correctly describe the conditional distribution of human feedback.
For example, when human preferences do not follow a
Bradley--Terry model, the target represents the best-fitting
Bradley--Terry scores under the fixed prompt distribution and
uniform weighting of model pairs.
This robustness is useful in LLM evaluation, where preferences
may vary across prompts in ways that a simple parametric model
cannot capture.
However, under misspecification, fitting the working model without
adjusting for unequal comparison probabilities gives greater
influence to frequently sampled pairs.
Different sampling policies can therefore lead to different
estimated scores and rankings, even with arbitrarily large samples
and unchanged underlying human preferences.
Hence we need to account for adaptive pair selection to estimate the fixed
population target, whose definition remains unchanged as the
comparison policy evolves.

For our main ranking application, we focus on the full score vector
$\bm\theta^\star$
and its induced ranks
\(
r_i^\star, i\in[K].
\)
We consider three inferential tasks:
constructing rank confidence sets,
certifying pairwise orderings $r_i^\star<r_j^\star$,
and certifying top-$k$ membership $r_i^\star\leq k$.
At a prespecified evaluation time $T$, we require the probability
that all rank sets cover their targets and all reported
certifications are correct to be asymptotically at least
$1-\alpha$.
Our aim is to use auxiliary information to obtain narrower rank
confidence sets and certify more ranking statements while
maintaining this simultaneous validity guarantee.

\section{Prediction-Powered Ranking Inference}

\subsection{Estimator}

Given adaptively collected human preferences $\{(\bm X_t,A_t,Y_t)\}_{t=1}^T$, unlabeled prompts $\{\tilde{\bm X}_i\}_{i=1}^N$, and an LLM judge $\bm f=(f_a)_{a\in\mathcal A}$, our goal is to estimate the evaluation target $\bm\theta^\star$ defined in \eqref{eqn:Z-estimation}. To motivate PP-Rank, first suppose that only human feedback
is available. A natural approach, following \citet{guo2025}, is to solve
the inverse-propensity-weighted empirical estimating equation:
\begin{equation}\label{eq: LabeledOnlyEstEqn}
    \frac{1}{T}\sum_{t=1}^T
    \frac{1}{\pi_t(A_t)}
    \bm g(\bm X_t,A_t,Y_t;\bm\theta)
    = \bm 0.
\end{equation}
The inverse propensity weights $1/\pi_t(a)$ compensate for unequal,
history-dependent comparison probabilities, so that the
estimating equation targets the population quantity
in \eqref{eqn:Z-estimation}.

To improve precision with limited human feedback, PP-Rank augments this equation with judge predictions on the auxiliary prompts. Because these predictions may be biased, their contribution is accompanied by a correction based on the observed human preferences. Let $\bm\lambda_T=(\lambda_{a,T})_{a\in\mathcal A}$ denote a vector of tuning weights that controls the
contribution of judge predictions for each model pair. We define the prediction-powered estimating function as
\begin{equation}\label{eq: esteqn}
\begin{aligned}
    \bm{G}_{T}^{\bm\lambda_{T}}(\bm{\theta})
    &= \frac{1}{N}\sum_{i=1}^N \sum_{a\in\mathcal{A}}
    \lambda_{a,T}\bm{g}(\tilde{\bm{X}}_i,a,f_a(\tilde{\bm{X}}_i);\bm{\theta}) \\
    &\quad + \frac{1}{T}\sum_{t=1}^{T}\frac{1}{\pi_t(A_t)}
    \bigg(\bm{g}(\bm{X}_t,A_t,Y_t;\bm{\theta})
    - \lambda_{A_t,T}\bm{g}(\bm{X}_t,A_t,f_{A_t}(\bm{X}_t);\bm{\theta})\bigg),
\end{aligned}
\end{equation}
and define the PP-Rank estimator $\widehat{\bm\theta}_T$ as an approximate solution to
$
    \bm G_T^{\bm\lambda_T}(\widehat{\bm\theta}_T)
    = \bm0.
$
Specifically, our asymptotic results allow a bounded estimator sequence satisfying
$   \bigl\|
    \bm G_T^{\bm\lambda_T}(\widehat{\bm\theta}_T)
    \bigr\|_2
    =o_p(T^{-1/2}).
$
This permits approximate root whose numerical error is negligible relative to the $T^{-1/2}$ statistical error.

The first term in $(\ref{eq: esteqn})$ aggregates judge-based score contributions across all model pairs and auxiliary prompts. The second corrects this prediction-based contribution using human feedback, with inverse propensity weights accounting for adaptive pair selection. The two prediction contributions cancel in expectation and $\mathbb E[\bm G_T^{\bm\lambda_T}(\bm\theta)]$ equals the population score in \eqref{eqn:Z-estimation}. Setting $\bm\lambda_T=\bm0$ recovers the human-only estimating equation \eqref{eq: LabeledOnlyEstEqn}. 

The weights $\bm\lambda_T$ control how strongly the estimator uses the LLM-judge predictions and are chosen to reduce variance. Setting $\lambda_{a,T}\approx 0$ largely discards the judge predictions for action $a$, which is useful when that prediction contributes little variance reduction. Tuning $\lambda_{a,T}$ can also reduce the variance caused by large inverse propensity weights, since $1/\pi_t(a)$ multiplies the residual $\bm{g}(\bm{X}_t,a,Y_t(a);\bm{\theta})-\lambda_{a,T}\bm{g}(\bm{X}_t,a,f_a(\bm{X}_t);\bm{\theta})$. When action $a$ is infrequently sampled, an appropriate choice of $\lambda_{a,T}$ can shrink this residual and reduce the effect of the large inverse propensity weight. Our experiments demonstrate that PP-Rank can maintain comparable inferential precision under substantially smaller exploration probabilities.

\subsection{Assumptions}

We now introduce the regularity conditions used in our theoretical analysis.
We assume that the environment generates contexts and potential outcomes independently over time:
$\{\bm X_t,Y_t(a):a\in\mathcal{A}\}_{t\geq 1} \stackrel{i.i.d.}{\sim} \mathcal{P}.$
This assumption is standard in contextual bandit inference \citep{zhang2021statistical,RuiSong, guo2025}.
However, since actions $A_t$ are selected adaptively, the labeled samples $\{(\bm X_t,Y_t):A_t=a\}$ need not be i.i.d., unlike the standard PPI setting.
In addition, we assume that the unlabeled prompts are independent of the labeled prompts and potential outcomes:
$\{\tilde{\bm X}_i\}_i \perp \{\bm X_t, \{Y_t(a)\}_{a\in\mathcal{A}}\}_{t\geq 1}.$

We also impose the following assumptions on the policy $\pi.$

\begin{assumption}[Policy Convergence]\label{as: polcon}
    There exists a stationary policy $\Bar{\pi}:\mathcal{X}\to\Delta(\mathcal{A})$ such that
    $\pi_t(a|\bm{X}_t,\mathcal{F}_{t-1})-\Bar{\pi}(a|\bm{X}_t) \xrightarrow[]{p} 0 \text{ as } t \to \infty.$ 
\end{assumption}

\begin{assumption}[Minimum sampling probability]\label{as: min prob}
    For $a \in \mathcal{A}$, there exists a constant $\pi_{\min}\in (0,1)$ such that $\pi_t(a) := \pi_t(a|\bm X_t, \mathcal{F}_{t-1}) \geq \pi_{\min}$ almost surely for all $t \geq 1$.
\end{assumption}

Policy convergence requires the probability of selecting each action, conditional on the current context, to stabilize asymptotically.
This ensures the policy's long-run behavior does not depend on early random fluctuations.
This assumption is satisfied by many commonly used adaptive policies and is standard in adaptive inference works under misspecification \citep{RuiSong, guo2025}.
Minimum sampling probabilities guarantee that no action is selected too infrequently.
This is particularly important under model misspecification, where persistent exploration helps ensure that the target can be estimated reliably.
Minimum sampling probabilities are also commonly assumed in the adaptive inference literature, especially under model misspecification \citep{zhang2021statistical,RuiSong,Zhang2022StatisticalIA32,guo2025, simchi2025multi, han2025}.
These assumptions are mild in LLM evaluation, since typically the platform controls the sampling policy and can design it to satisfy both conditions.

\begin{assumption}[Unconfoundedness]\label{as: unconfounded}
    $A_t \perp \{Y_t(a)\}_{a\in\mathcal{A}}|(\mathcal{F}_{t-1},\bm{X}_t)$ for $t=1,\dots,T.$
\end{assumption}

Assumption \ref{as: unconfounded} formalizes the standard idea that the decision-maker acts without knowledge of the current potential outcomes beyond the information already observed. It holds, for example, when an evaluation platform assigns a model pair before observing the corresponding human preference.
Analogous assumptions are common in the contextual bandit inference literature \citep{zhang2021statistical, guo2025,han2025}.

When the inferential target is Bradley--Terry scores, we impose
the following assumption. Analogous conditions for general targets are given in Appendix~S2.

\begin{assumption}[Nondegenerate win probabilities]
\label{as: win-prob}
    $\forall \text{ } (i,j)\in\mathcal{A}$, $\mathbb{P}\{Y(i,j)=1\}\in(0,1).$
\end{assumption}

Assumption~\ref{as: win-prob} requires that neither model wins a
pairwise comparison with probability one; this ensures that the Bradley--Terry scores are finite.

\subsection{Estimation Theory}

We now state our main theoretical results.
Let $r = \lim_{T,N\to\infty}T/N$ denote the asymptotic ratio of the number of labeled and unlabeled observations.
Since unlabeled tasks are often far more abundant than human-labeled tasks, $r$ is typically small.
For any $\bm v \in \mathbb R^d,$ define $\bm v^{\otimes 2} := \bm v\bm v^\top.$
The following result characterizes the asymptotic distribution of the estimator.
We state the result for Bradley--Terry scores;
Appendix~S1.1 establishes
the same conclusion for general estimating functions
under regularity conditions given in
Appendix~S2.
\begin{theorem}[Asymptotic Normality] \label{thm: asymptotic}
    Consider the Bradley--Terry target defined by
\eqref{eqn:Z-estimation} with estimating function
\eqref{eqn:BTL_g}.
Suppose Assumptions~\ref{as: polcon}--\ref{as: win-prob}
hold, and suppose
$\bm\lambda_T \xrightarrow{p} \bar{\bm\lambda}$.
    Then there exists an estimator sequence $\{\widehat{\bm{\theta}}_{T}\}_{T\geq 1}$ satisfying $\bigl\|
    \bm G_T^{\bm\lambda_T}(\widehat{\bm\theta}_T)
    \bigr\|_2
    =o_p(T^{-1/2})$ with $||\widehat{\bm{\theta}}_{T}||_2 \leq R_{\theta}$ for all $T.$
    Moreover, for any such sequence, as $T \to \infty,$
    \begin{equation}\label{eq: asymnorm}
        \sqrt{T}(\widehat{\bm{\theta}}_{T}-\bm{\theta}^*) \xrightarrow[]{d} \mathcal{N}(\bm{0},\bm\Sigma_{\overline{\lambda}}),
    \end{equation}
    where 
    \begin{equation}\label{eq: sigma}
        \bm\Sigma_{\overline{\bm\lambda}}=\bm J^{-1}(r\mathrm{Cov}(\bm W)+\bm V)\bm J^{-\top},
    \end{equation}
    $${\bm J} = \sum_{a \in \mathcal{A}} \mathbb E[\nabla\bm g(\bm X,a,Y(a);\bm\theta^*)],\quad\quad \bm W = \sum_{a \in \mathcal{A}}\bar\lambda_a\bm g(\bm X,a,f_a(\bm X);\bm\theta^*),$$ 
    $$\bm V = \sum_{a \in \mathcal{A}} \mathbb E\left[\frac{1}{\bar\pi(a)}\left(\bm g(\bm X,a,Y(a);\bm\theta^*)-\bar\lambda_a\bm g(\bm X,a,f_a(\bm X);\bm\theta^*)\right)^{\otimes 2}\right] - \mathbb E[\bm W]^{\otimes 2}.$$
\end{theorem}

The asymptotic variance $(\ref{eq: sigma})$ separates uncertainty from the  auxiliary sample, $r\operatorname{Cov}(\bm W)$, from the human correction, $\bm V$. The latter contains the residual score weighted by inverse sampling probabilities. An informative judge can reduce residual variation, particularly for infrequently sampled pairs, while the term $r\operatorname{Cov}(\bm W)$ records the uncertainty of its auxiliary-sample average. When $N/T\to\infty$, that auxiliary contribution vanishes.

Establishing Theorem~\ref{thm: asymptotic} presents several
challenges beyond standard prediction-powered inference
\citep{PPI,PPI++}. First, because human comparisons are selected adaptively, the observed feedback is not i.i.d., and the sampling policy may also depend on the auxiliary data. 
These features must be handled while allowing the working
preference model to be misspecified. To address these challenges, our proof combines a uniform law of
large numbers and Z-estimator linearization with a joint martingale central limit theorem. By incorporating the auxiliary observations and adaptively collected human feedback into a common martingale array, we establish
asymptotic normality while accounting for their dependence.
A further challenge arises because the prediction weights
$\bm\lambda_T$ are estimated from the same data used for inference. We show that consistent estimation of $\bm\lambda_T$ introduces only an $o_p(T^{-1/2})$ error and does not affect the first-order asymptotic distribution. The detailed proof is provided in Appendix~S1.1.

The asymptotic normality in \eqref{eq: asymnorm} holds for any limiting weight vector $\bar{\bm\lambda}$, which controls how strongly LLM-judge predictions contribute.
However, its choice affects the estimator's asymptotic variance.
The following theorem characterizes the optimal weights $\bm\lambda^*$ that minimize $\operatorname{tr}(\bm\Sigma_{\bar{\bm\lambda}})$, thereby making the most efficient use of LLM-judge predictions.

\begin{theorem}[Optimal $\bar{\bm\lambda}$]\label{thm: lambda}
Assuming $\bm Q$ is positive definite, the trace of $\bm\Sigma_{\bar{\bm\lambda}}$ is minimized at
\vspace{-1em}
\begin{equation}\label{eq: lambdastar}
\bm\lambda^*=\bm Q^{-1}\bm b,
\vspace{-1em}
\end{equation}
where, for $a,a'\in\mathcal A$,
\[
\begin{aligned}
\bm Q_{a,a}
&=
\tr\!\left[
\bm J^{-1}
\left(
\mathbb E\!\left[
\left(r+\frac{1}{\bar\pi(a\mid\bm X)}\right)
\bm g_{f,a}^{\otimes 2}
\right]
-(r+1)\mathbb E[\bm g_{f,a}]^{\otimes 2}
\right)
\bm J^{-1\top}
\right],
\\[0.4em]
\bm Q_{a,a'}
&=
\tr\!\left[
\bm J^{-1}
\left(
r\,\mathbb E[\bm g_{f,a}\bm g_{f,a'}^\top]
-(r+1)\mathbb E[\bm g_{f,a}]
\mathbb E[\bm g_{f,a'}]^\top
\right)
\bm J^{-1\top}
\right],
\qquad a\neq a',
\\[0.4em]
\bm b_a
&=
\tr\!\left[
\bm J^{-1}
\mathbb E\!\left[
\frac{\bm g_{Y,a}\bm g_{f,a}^\top}
{\bar\pi(a\mid\bm X)}
\right]
\bm J^{-1\top}
\right].
\end{aligned}
\]
Here,
$\bm g_{f,a}=\bm g(\bm X,a,f_a(\bm X);\bm\theta^*)$ and
$\bm g_{Y,a}=\bm g(\bm X,a,Y(a);\bm\theta^*)$.
\end{theorem}

Since $\bm\lambda^*$ depends on population quantities, it must be estimated with observed data.
We provide a plug-in estimator $\hat{\bm\lambda}_T$ in \textup{(S20)} of Appendix S3.2.
Using $\hat{\bm\lambda}_T$ in the PP-Rank estimator minimizes the trace of its asymptotic variance, as the following corollary establishes.

\begin{corollary}\label{cor: minimalvar}

    Under the conditions of Theorem~\ref{thm: asymptotic} and Theorem~S1, let $\bm\lambda_T=\widehat{\bm\lambda}_T$ from \textup{(S20)}.
    Then the asymptotic variance of $\widehat{\bm\theta}_{T}$ has minimal trace among all variances of the form \eqref{eq: sigma}.
\end{corollary}

Since $\bm\lambda = \bm 0$ corresponds to discarding the LLM-judge feedback, Corollary \ref{cor: minimalvar} implies that our estimator incorporates the predictions and unlabeled data without increasing the trace of its asymptotic variance. Moreover, unlike standard PPI power tuning, which produces a single weight for the LLM judge, Theorem \ref{thm: lambda}  yields a vector of action-specific weights. This makes the estimator robust to two challenges that may vary across model pairs: inaccurate LLM-judge feedback and large inverse propensity weights. When the LLM-judge feedback is inaccurate for a particular model pair $a=(i,j)$, the weight $\lambda_a^*$ can be small, limiting the influence of its predictions. Consistent with this behavior, the simulations in Appendix~S5.3 show that $\widehat{\bm\lambda}_T$, averaged over model pairs, approaches one for a perfect judge and zero for an uninformative judge. Alternatively, if $a$ is selected infrequently, $\lambda_a^*$ can be chosen to reduce the prediction residual
$\bm g(\bm X_t,a,Y_t;\bm\theta)-\lambda_a\bm g(\bm X_t,a,f_a(\bm X_t);\bm\theta)$
and thereby mitigate the variance induced by inverse propensity weighting.

\subsection{Simultaneous Inference for Ranking}\label{sec: ranking}

We now apply the asymptotic theory to simultaneous inference
on LLM rankings. For the Bradley--Terry target introduced
in Section~3, recall that
$\bm\theta^\star\in\mathbb R^{K-1}$ contains the free
score parameters, while
$\bm\beta^\star=(\beta_1^\star,\ldots,\beta_K^\star)^\top$
denotes the full vector of model scores.
Define
\(
\bm B=(\bm I_{K-1},-\bm 1_{K-1})^\top
\in\mathbb R^{K\times(K-1)},
\)
so that
$\bm\beta^\star=\bm B\bm\theta^\star$
and
$\widehat{\bm\beta}_T=\bm B\widehat{\bm\theta}_T$. By Theorem~\ref{thm: asymptotic}, we have
\(
\sqrt T
(\widehat{\bm\beta}_T-\bm\beta^\star)
\xrightarrow{d}
\mathcal N(\bm0,\bm\Omega_{\bar{\bm\lambda}}),
\)
where
\(
\bm\Omega_{\bar{\bm\lambda}}
=
\bm B\bm\Sigma_{\bar{\bm\lambda}}\bm B^\top
\in\mathbb R^{K\times K},
\)
which establishes joint asymptotic distribution of the model scores of all $K$ models. To estimate this covariance matrix, let
$\widehat{\bm\Sigma}_T$ denote the plug-in estimator
of $\bm\Sigma_{\bar{\bm\lambda}}$ constructed in
Appendix~S3.3.
Proposition~S1 of Appendix~S3.3
establishes its consistency.
We then estimate the covariance matrix of the full
model scores by
\(
\widehat{\bm\Omega}_T
=
\bm B\widehat{\bm\Sigma}_T\bm B^\top.
\)
Consequently,
$\widehat{\bm\Omega}_T
\xrightarrow{p}
\bm\Omega_{\bar{\bm\lambda}}$. Now we are ready to conduct simultaneous inference and construct simultaneous rank confidence sets.

For each $i\neq j$, define the pairwise score difference
\(
\Delta_{ij}^\star
=
\beta_i^\star-\beta_j^\star,
\widehat\Delta_{ij}
=
\widehat\beta_{i,T}-\widehat\beta_{j,T},
\)
and let
$\bm c_{ij}=\bm e_i-\bm e_j\in\mathbb R^K$,
where $\bm e_i$ denotes the $i$th standard basis vector.
Its estimated and population asymptotic variances are, 
\(
\widehat s_{ij}^2
=
\bm c_{ij}^\top
\widehat{\bm\Omega}_T
\bm c_{ij},
\)
and
\(
s_{ij}^2
=
\bm c_{ij}^\top
\bm\Omega_{\bar{\bm\lambda}}
\bm c_{ij},
\)
respectively.
We assume $s_{ij}^2>0$ for every $i\neq j$. Let $\widehat c_{1-\alpha}$ be the conditional
$(1-\alpha)$-quantile of
\(
\max_{i<j}
\frac{|\bm c_{ij}^\top\bm Z|}{\widehat s_{ij}},
\bm Z\sim
\mathcal N(\bm0,\widehat{\bm\Omega}_T).
\)
We construct the simultaneous confidence intervals
\[
\mathcal I_{ij}
=
[L_{ij},U_{ij}]
=
\left[
\widehat\Delta_{ij}
-
\frac{\widehat c_{1-\alpha}\widehat s_{ij}}{\sqrt T},
\;
\widehat\Delta_{ij}
+
\frac{\widehat c_{1-\alpha}\widehat s_{ij}}{\sqrt T}
\right],
\qquad i<j.
\]
We certify that model $i$ outperforms model $j$  whenever $L_{ij}>0$.

\begin{corollary}[Simultaneous pairwise inference]
\label{cor: simultaneous-pairwise}
Under the conditions of Theorem~\ref{thm: asymptotic}
and Proposition~S1 of the supplement,
suppose $s_{ij}^2>0$ for every $i<j$.
Then as $T, N \to \infty,$
\[
\mathbb P\left(
\Delta_{ij}^\star\in\mathcal I_{ij}
\text{ for all }i<j
\right)
\longrightarrow 1-\alpha.
\]
\end{corollary}

Unlike marginal inference for individual model scores,
simultaneous inference must account for dependence among
all $\binom K2$ pairwise comparisons.
These contrasts share estimated model scores and are
generally correlated, with their dependence further
affected by adaptive sampling and the use of auxiliary
judge predictions. Corollary~\ref{cor: simultaneous-pairwise} addresses this
issue by studying the maximum of the standardized
pairwise contrasts using the estimated  covariance.
Its proof follows from the joint Gaussian limit in
Theorem~\ref{thm: asymptotic}, consistency of
$\widehat{\bm\Omega}_T$, and the continuous mapping
and Slutsky theorems.

We next translate these simultaneous confidence intervals
into uncertainty quantification for the induced ranking.
Recall that the rank of model $i$ is 
\(
r_i^\star
=
1+
\sum_{j\neq i}
\mathbf 1\{\beta_j^\star>\beta_i^\star\}.
\)
The number of models certified to outperform model $i$
provides a lower bound on its rank, while the number
certified to be worse provides an upper bound.
Accordingly, define
\[
\widehat r_i^L
=
1+
\sum_{j\neq i}
\mathbf 1\{U_{ij}<0\},
\qquad
\widehat r_i^U
=
K-
\sum_{j\neq i}
\mathbf 1\{L_{ij}>0\}.
\]
The rank confidence set for model $i$ is
\(
\mathcal R_i
=
\{
\widehat r_i^L,
\widehat r_i^L+1,
\ldots,
\widehat r_i^U
\}.
\)

\begin{corollary}[Simultaneous rank confidence sets]
\label{cor: simultaneous-ranks}
Under the conditions of
Corollary~\ref{cor: simultaneous-pairwise},
\[
\mathbb P\left(
r_i^\star\in\mathcal R_i
\text{ for all }i\in[K]
\right)
\geq
1-\alpha+o(1).
\]
\end{corollary}

The proof follows from a deterministic implication:
whenever all pairwise confidence intervals cover their
true score differences, every certified pairwise ordering
is correct, which
implies simultaneous coverage of all rank confidence sets.

For $k\in\{1,\ldots,K-1\}$, define the population
top-$k$ set as
\(
\mathcal T_k^\star
=
\{i:r_i^\star\leq k\}.
\)
We certify that a model belongs to the top-$k$
if its rank upper bound is at most $k$, and a model falls outside
the top-$k$ if its rank lower bound exceeds $k$.
The resulting certification sets are
\[
\widehat{\mathcal T}_k^{\mathrm{in}}
=
\{i:\widehat r_i^U\leq k\},
\qquad
\widehat{\mathcal T}_k^{\mathrm{out}}
=
\{i:\widehat r_i^L>k\}.
\]

\begin{corollary}[Top-$k$ certification]
\label{cor: top-k}
Under the conditions of
Corollary~\ref{cor: simultaneous-pairwise},
for any $k$,
\[
\mathbb P\left(
\widehat{\mathcal T}_k^{\mathrm{in}}
\subseteq
\mathcal T_k^\star
\quad\text{and}\quad
\widehat{\mathcal T}_k^{\mathrm{out}}
\subseteq
[K]\setminus\mathcal T_k^\star
\right)
\geq
1-\alpha+o(1).
\]
\end{corollary}

Corollary~\ref{cor: top-k} provides a practical guarantee
for LLM leaderboard decisions: with asymptotic probability
at least $1-\alpha$, every model certified to belong
to the top-$k$ is indeed in that set, and every model
certified to fall outside it is correctly excluded. Importantly, the precision gains from incorporating
LLM-judge predictions can translate into narrower
rank confidence sets and more frequent certification
of ranking statements. Our experiments demonstrate
improvements in rank-set precision and certification
relative to the baseline.

\section{Sampling Requirements and Evaluation Costs}\label{sec: costreduc}

In evaluation settings, decision-makers must account for the costs or detrimental effects associated with certain actions. For example, an AI lab may wish to limit the number of completions generated by an expensive model, while a ranking platform may wish to avoid showing users weak models that could reduce user satisfaction. In this section, we study when auxiliary LLM-judge feedback can reduce the exploration required for reliable inference and thereby allow costly actions to be selected less frequently.

\begin{remark}[Inference with a perfect judge]
Suppose that, for every $a\in\mathcal A$ and $\bm\theta\in\Theta$, $\bm g(\bm X,a,Y(a);\bm\theta)=\bm g(\bm X,a,f_a(\bm X);\bm\theta)$ a.s., and set $\bm\lambda_T\equiv\bm 1$. Then the PP-Rank estimating equation \eqref{eq: esteqn} is
\[
\bm G_T^{\bm 1}(\bm\theta)
=
\frac{1}{N}
\sum_{i=1}^{N}
\sum_{a\in\mathcal A}
\bm g\!\left(
\widetilde{\bm X}_i,
a,
f_a(\widetilde{\bm X}_i);
\bm\theta
\right),
\]
which is independent of the sampling policy $\pi$. In particular, valid inference remains possible even when the sampling probability $\pi_t(a)$ converges to zero.
\end{remark}

The preceding remark describes an idealized setting in which a perfect LLM judge eliminates the need for human feedback. In practice, however, human preferences are inherently noisy, and even a highly accurate judge cannot perfectly reproduce individual human judgments. Consequently, continued human sampling remains important for reliable inference. The following theorem formalizes this limitation by showing that vanishing sampling probabilities preclude regular $\sqrt{T}$-rate estimation of pairwise win probabilities.

\begin{theorem}[Necessity of persistent human sampling]
\label{thm: persistent_sampling}
Suppose that, for some $a\in\mathcal A$,
$\mathbb P(Y_t(a)=1\mid\bm X_t)=p_a$ a.s.,
where $p_a\in(0,1)$.
If $\pi_t(a)\xrightarrow{p}0$, then no estimator of $p_a$
is regular at the $\sqrt T$ rate.
\end{theorem}
Regularity is a standard condition requiring the estimator to behave stably under small perturbations of the data-generating distribution. Henceforth, we assume that human preferences are noisy, that is, 
for a general evaluation target,
\(
\mathbb E\left[
\operatorname{tr}\left\{
\operatorname{Var}\big(
\bm g(\bm X,a,Y(a);\bm\theta^\star)
\mid \bm X
\big)
\right\}
\right]>0, \forall a\in\mathcal A.
\) Theorem~\ref{thm: persistent_sampling} shows that when the probability of selecting a model pair vanishes,
regular $\sqrt T$-rate estimation of its  win probability becomes impossible, even with access to a fixed external LLM judge.
This result does not rule out consistent estimation at slower rates, but highlights the statistical cost of insufficient exploration. %
We next develop an online procedure that learns how far the random sampling of costly actions can be reduced without sacrificing inferential precision.

A common way to balance inferential precision and action cost is to combine a low-cost sampling rule with uniform exploration. Let \(\rho(\cdot\mid \bm X_t,\mathcal{F}_{t-1})\) denote a policy that favors less costly actions. We consider policies of the form
\[ \pi_t(a\mid X_t,\mathcal{F}_{t-1}) = \pi_{\min} + (1-|\mathcal A|\pi_{\min}) \rho(a\mid X_t,\mathcal{F}_{t-1}). \]
The parameter \(\pi_{\min}\) controls the tradeoff between statistical precision and cost. 
A larger $\pi_{\min}$ raises the guaranteed minimum sampling probability across model pairs, thereby tightening the conservative upper bound on the variance induced by inverse-propensity weighting. However, it also increases forced exploration and reduces the probability allocated to comparisons favored by $\rho$.
Conversely, a smaller \(\pi_{\min}\) allows the policy to follow the low-cost rule \(\rho\) more aggressively, at the expense of greater inferential variance.

Suppose the labeled-only estimator is used with a baseline exploration rate \(\pi_{\min}^{\mathrm{base}}\). We treat the resulting asymptotic variance as the benchmark level of inferential precision. Because PP-Rank uses LLM-judge predictions to reduce variance, it may achieve the same precision with a smaller exploration rate. We therefore seek some \(\pi_{\min}^{PP}\le \pi_{\min}^{\mathrm{base}}\) such that PP-Rank is no less precise than the labeled-only baseline:
\begin{equation}\label{eq: variance_preservation}
\operatorname{tr}\!\left\{
\bm\Sigma_{\lambda^\star(\pi_{\min}^{PP})}
(\pi_{\min}^{PP})
\right\}
\le
\operatorname{tr}\!\left\{
\bm\Sigma_0(\pi_{\min}^{\mathrm{base}})
\right\}.
\end{equation}
In this way, the efficiency gain provided by the LLM judge is converted into reduced exploration, and lower evaluation cost, while maintaining the same level of inferential precision.

To choose \(\pi_{\min}^{PP}\), we require a variance criterion that depends only on the minimum sampling probability rather than on the full adaptive policy. For any candidate minimum sampling probability $p>0$, we define $\bm\Sigma_{\bm\lambda}(p)$ as,
\begin{equation*}
\begin{aligned}
\bm\Sigma_{\bm\lambda}(p)
=\bm J^{-1}\Bigg[
r\,\operatorname{Cov}\Big(\sum_{a\in\mathcal A}\lambda_a\bm g_{f,a}\Big)
+\frac{1}{p}\sum_{a\in\mathcal A}
\mathbb E\big[(\bm g_{Y,a}-\lambda_a\bm g_{f,a})^{\otimes 2}\big]
-
\Big(\sum_{a\in\mathcal A}\lambda_a\mathbb E[\bm g_{f,a}]\Big)^{\otimes 2}
\Bigg]\bm J^{-\top},
\end{aligned}
\end{equation*}
where $\bm g_{f,a}$ and $\bm g_{Y,a}$ were defined in Theorem \ref{thm: lambda}.
This quantity is obtained from the asymptotic variance formula \eqref{eq: sigma} by replacing each action propensity with the lower bound $p$. Thus, $\bm\Sigma_{\bm\lambda}(p)$ represents a worst-case upper bound on the asymptotic variance over policies with minimum sampling probability $p$, and \eqref{eq: variance_preservation} requires that reducing the minimum sampling probability does not increase this worst-case variance relative to the labeled-only baseline. For each $p$, we define
$\bm\lambda^*(p)
=
\underset{\bm\lambda\in\mathbb R^{|\mathcal A|}}
{\operatorname{argmin}}\;
\tr\!\left(
\bm\Sigma_{\bm\lambda}(p)
\right),$
analogously to \eqref{eq: lambdastar}. We choose $\pi_{\min}^{\mathrm{PP}}$ as the solution to
\begin{equation}\label{def: eps_equiv}
\tr\!\left(
\bm\Sigma_{\bm\lambda^*(\pi_{\min}^{\mathrm{base}})}
(\pi_{\min}^{\mathrm{PP}})
\right)
=
\tr\!\left(
\bm\Sigma_{\bm0}(\pi_{\min}^{\mathrm{base}})
\right).
\end{equation}

\begin{proposition}\label{prop: baseline_matching}
Suppose Assumptions~\ref{as: polcon}--\ref{as: unconfounded}
of the main text and
Assumptions~S1--S3
of Appendix~S2 hold.
Let $0<\pi_{\min}^{\mathrm{base}}\leq 1/|\mathcal A|$,
and assume that the matrix $\bm Q$ in
Theorem~\ref{thm: lambda}, with $\bar\pi(a\mid\bm X)$
replaced by $\pi_{\min}^{\mathrm{base}}$, is invertible.
Assume $T/N \to r$ and $\mathbb E[\tr(\operatorname{Var}(\bm g(\bm X,a,Y(a);\bm\theta^*)\mid\bm X))]>0$ for all $a\in\mathcal A$. Then \eqref{def: eps_equiv} has a unique solution, which satisfies $\pi_{\min}^{\mathrm{PP}}\leq\pi_{\min}^{\mathrm{base}}$. 
\end{proposition}

In practice, the quantities in \eqref{def: eps_equiv} are unknown
and must be estimated. We provide a consistent estimator
$\widehat{\pi}_{\min,t}^{\mathrm{PP}}$ in
Appendix~S4.3.

\begin{proposition}\label{prop: pimin-consistency}
Under the conditions of Proposition~\ref{prop: baseline_matching},
let $\hat{\bm\theta}^{\mathrm{label}}_T$ solve the
labeled-only estimating equation~\eqref{eq: LabeledOnlyEstEqn} and $\hat{\bm\theta}^{\mathrm{label}}_T\xrightarrow{p}\bm\theta^*.$ Assume $T/N\to r.$
Then as $t\to\infty$, $\hat{\pi}_{\min,t}^{\mathrm{PP}}\xrightarrow{p}\pi_{\min}^{\mathrm{PP}}$. 
\end{proposition}
The decision-maker can update the minimum sampling probability sequentially using
$\pi_{\min,t}
=
\max\left\{
\widehat{\pi}_{\min,t-1}^{\mathrm{PP}},
\underline{\pi}
\right\},$
where the clip
$\underline{\pi}\in(0,1/|\mathcal{A}|]$
prevents any model pair from being sampled too infrequently and can be very small in practice. The decision-maker then employs
$
\pi_t(a\mid\bm X_t,\mathcal H_{t-1})
=
\pi_{\min,t}
+
\left(1-|\mathcal{A}|\pi_{\min,t}\right)
\rho(a\mid\bm X_t,\mathcal H_{t-1}).$
Experiments in Section~\ref{sec: synthetic} show that this procedure can substantially reduce evaluation costs relative to the labeled-only baseline without sacrificing inferential precision.

\section{Semi-Synthetic LLM Ranking Simulation}\label{sec: semisynthetic}

We evaluate our method in a semi-synthetic experiment modeled on LLM ranking platforms.
We first describe the prompts, response models, LLM judges, simulated user preferences, and adaptive model-selection policy; then we present the results.

We use the UltraFeedback-Extended dataset \citep{cui2024ultrafeedback} as our source of prompts, model responses, and LLM-judge scores.
The dataset contains approximately 64,000 prompts, responses from 10 LLMs to each prompt, and scores from 10 LLM judges for each response.
The prompts cover diverse tasks and domains, resembling the
heterogeneous prompt distribution encountered in LLM ranking
platforms. We rank the 10 response models and combine the
10 component judges into a single ensemble judge.
Appendix~S5.1 lists the models and
prompt sources.

In our experiments, we construct an ensemble judge by averaging the scores assigned by multiple judges.
For a prompt $\bm X$ and a pair of response models $(i,j) \in [10]^2,$ let $R_i$ and $R_j$ denote the scores assigned by the judge to their respective responses.
We define the judge's predicted preference for model $i$ over model $j$ as $f_{(i,j)}(\bm X) := \sigma\left(\frac{R_i-R_j}{\tau}\right),$
where $\sigma(\cdot)$ is the sigmoid function.
We hold out $20\%$ of prompts and choose $\tau$ to maximize the correlation between judge and simulated user preferences on that held-out set, giving $\tau = 0.4987$; these prompts are then excluded from the simulation.

Because UltraFeedback-Extended does not include human pairwise preferences, we simulate them using the reward model Skywork-Reward-V2-8B, which assigns a numerical score to each response.
When comparing two responses, we define the simulated user's preference to be the response with the larger reward-model score, unless the scores differ by less than a pre-selected $\delta > 0$, in which case the comparison is recorded as a tie ($Y=0.5$).
We choose $\delta$ such that approximately $27.5\%$ of the simulated preferences are ties, roughly matching Arena's empirical tie rate.

On LLM ranking platforms, the platform observes a prompt $\bm X$ and adaptively selects a pair of models $A=(i,j)$ whose responses are shown to the user.
We emulate this process using an adaptive pair-selection policy, which we now describe.
The policy begins with an adaptive burn-in phase, during which it selects uniformly among pairs whose comparisons would connect previously disconnected components of the current directed win graph.
This burn-in continues until the directed win graph is strongly connected; this ensures there are enough data to form an initial estimate of the Bradley-Terry scores.

After burn-in, we use an $\epsilon$-greedy policy with $\epsilon=0.1;$ specifically, with probability $\epsilon$ the policy selects a model pair uniformly at random.
Otherwise, it selects a pair uniformly from among the five models with the largest current estimated Bradley-Terry scores.
In this way, the policy concentrates comparisons on models believed to be near the top of the leaderboard while continuing to explore all pairs.

Unless otherwise noted, we conduct 1,000 simulation trials for each estimator. The best-model identification experiment in Table~\ref{tab: semiAdaptivity} uses 200 matched trials per arm.
Each trial contains $T=10{,}000$ battles with simulated human preferences and $N=30{,}000$ prompts for which only the LLM-judge preferences are used.
We employ a single scalar tuning parameter $\lambda$ shared across model pairs.
We compute the target BT vector $\bm\theta^*$ by fitting the BT model on all comparisons between all pairs of responses on all prompts.

At the end of each trial, we estimate the Bradley-Terry scores and ranks of all 10 models.
We evaluate the resulting rank confidence sets and test whether they certify conclusions of practical interest, such as whether a given model belongs to the top $k$.

\subsection{Results}

\paragraph{Benefit of adaptive sampling}
A preliminary experiment illustrates the inferential benefits of adaptive sampling when the goal is to identify the single best model. We compare uniform sampling with a policy targeted at this question. After each battle, the targeted policy recomputes the labeled-only IPW rank sets. With probability $0.9$, it selects uniformly among pairs involving at least one model whose top-1 membership is still uncertain; otherwise, it selects a pair uniformly at random. Once every model's top-1 membership is certified, the policy samples uniformly. Table~\ref{tab: semiAdaptivity} reports how often the top-1 set is correctly certified, meaning that the best model is certified in and the other nine models are certified out. With the labeled-only IPW estimator, adaptive sampling increases this rate from $38\%$ to $52\%$. Under the same sampling policy, PP-Rank raises the certification rate to $74\%$.

\begin{table}[!ht]
\centering
\begin{tabular}{lccc}
\hline
 & Uniform & Targeted & PP-Rank\\
\hline
Top-1 set certified & 0.38 & 0.52 & 0.74 \\
\hline
\end{tabular}
\caption{Proportion of trials in which the top-1 set is correctly certified. Uniform and Targeted use the labeled-only IPW estimator under uniform sampling and an adaptive policy targeted at the top-1 question, respectively. PP-Rank uses the same adaptive policy. We set $T=10{,}000$ and run 200 matched trials per arm.}
\label{tab: semiAdaptivity}
\end{table}

\paragraph{Validity under adaptive sampling}
To demonstrate the effect of adaptive sampling on validity and precision, we vary the adaptivity of the sampling policy. We compare against PPI++ \citep{PPI++}, which is also used in \cite{PPRanking}; a judge-only estimator that uses only LLM-judge preferences; and ``labeled-only'' estimators that use only human feedback, with and without inverse propensity weighting.

\begin{figure}[ht]
\centering{\includegraphics[width=0.9\columnwidth]{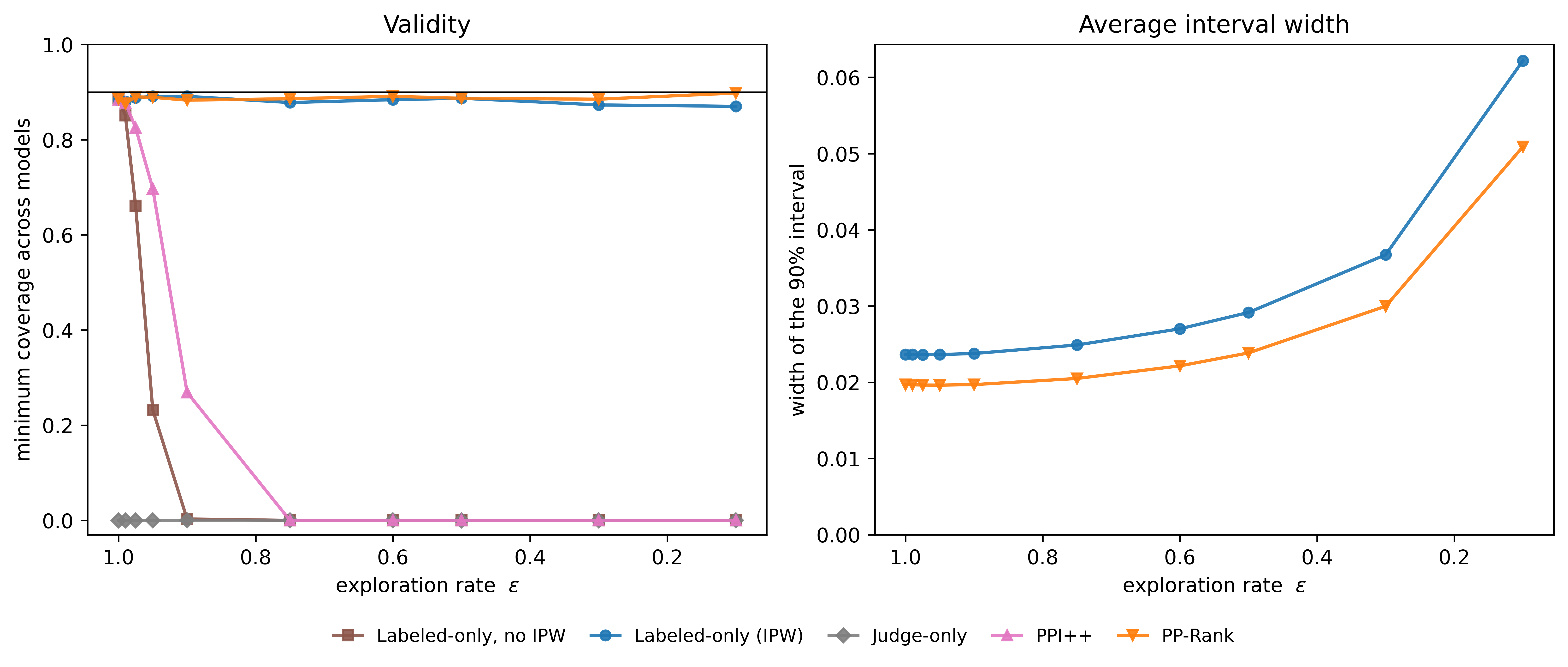}}
    \caption{Left: minimum coverage across the ten models' win rates against a uniformly chosen opponent, at nominal 90\%. Right: average width of the same intervals for the two estimators that remain valid. For each method we run 1000 trials with $T = 10{,}000$. Only the inverse-propensity-weighted estimators remain valid as exploration decreases, and PP-Rank's intervals are narrower at every exploration rate.}
    \label{fig: validitySemi}
\end{figure}

For each estimator, we construct nominal 90\% confidence intervals for each model's win rate and report the minimum empirical coverage across the ten models. At $\epsilon=1$, the sampling policy is uniform; as $\epsilon$ decreases, it becomes increasingly adaptive. The left panel of Figure \ref{fig: validitySemi} shows that PP-Rank and the labeled-only IPW baseline hold coverage near nominal at every exploration rate. It also shows that coverage deteriorates with increasing adaptivity for methods that do not use inverse propensity weighting.

\paragraph{PP-Rank improves precision and certification}
Among the two methods that remain valid, the right panel of Figure \ref{fig: validitySemi} shows that PP-Rank produces narrower intervals than the labeled-only IPW baseline at every exploration rate. Similarly, Table~\ref{tab: semiImprovement} reports PP-Rank's improvement in precision at
$\epsilon=0.1$ in estimating Bradley-Terry scores and in recovering the exact ranking (the
ranking induced by the estimated scores matching the true ranking).

\begin{table}[!ht]
\centering
\begin{tabular}{lccc}
\hline
 & Labeled-only & PP-Rank & Improvement (\%) \\
\hline
MSE ($\downarrow$) & 0.0112 & 0.0072 & 35.6\% \\
Avg.\ CI width ($\downarrow$) & 0.341 & 0.276 & 19.0\% \\
Avg.\ rank-set width ($\downarrow$) & 2.074 & 1.867 & 10.0\% \\
Exact ranking recovered ($\uparrow$) & 0.286 & 0.357 & 24.8\% \\
\hline
\end{tabular}
\caption{Comparison of PP-Rank with the labeled-only IPW baseline.
$T=10{,}000$, $\epsilon=0.1$, 1,000 trials.
Arrows indicate whether smaller ($\downarrow$) or larger ($\uparrow$) values are better.
Improvement is the relative percentage improvement compared with the labeled-only baseline.}
\label{tab: semiImprovement}
\end{table}

We also report certification of statements of the form ``model $m$ belongs to the top (or bottom) $K$''.
We focus on the 10 such statements that are of moderate difficulty in the sense that the labeled-only baseline certifies them in between \(10\%\) and \(90\%\) of trials.
Averaged over these, the certification rate rises from $0.527$ to $0.694$.
For example, the labeled-only baseline certifies that Yi-6B is among the three weakest models in $65.5\%$ of trials, and PP-Rank increases this rate to $88.5\%$.
Similarly, the labeled-only baseline certifies that Mixtral outperforms Ministral in $50.5\%$ of trials, and PP-Rank increases this rate to $66.2\%$.

\begin{figure}[ht]
    \centering
    \includegraphics[width=0.9\linewidth]{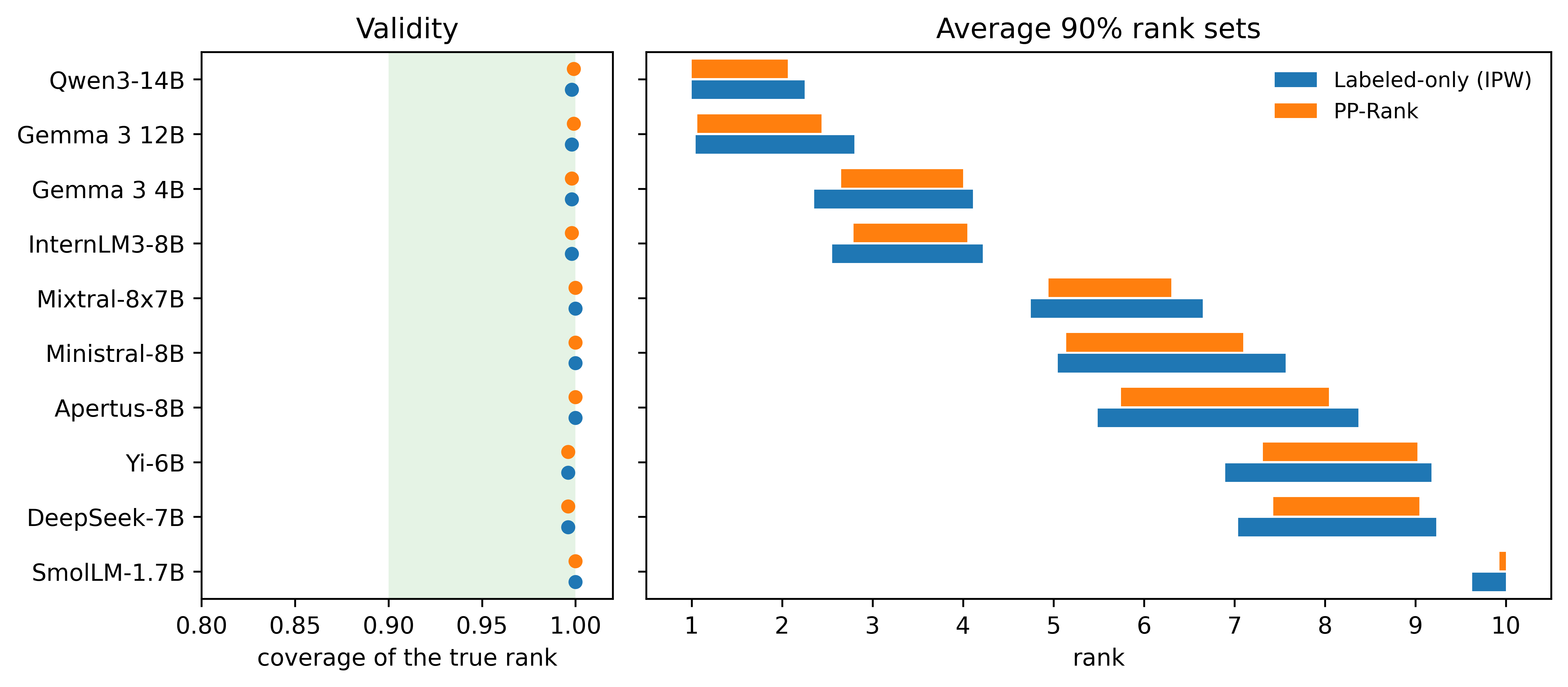}
    \caption{Left: percent of trials in which each model's rank set contains its true rank; the shaded region marks coverage of at least the nominal 90\%. Right: average 90\% rank sets for each model. $T = 4{,}000$, $\epsilon = 0.1$, 1,000 trials. Both methods achieve nominal coverage, and PP-Rank's rank sets are narrower.}
    \label{fig: ranksets}
\end{figure}

 Figure \ref{fig: ranksets} shows that PP-Rank produces rank sets with nominal coverage that are narrower than those of the labeled-only baseline.

\paragraph{Robustness to sampling policy}
Finally, we vary the sampling policy to show that the same qualitative behavior persists under alternative policy choices.
Under four policies, we construct confidence intervals for all 45 pairwise Bradley-Terry score contrasts and compare their minimum coverage. In addition to the greedy policy used above, we consider three alternatives. For a pair $a=(i,j)$, \textsc{Uncertainty} samples in proportion to
$w_t(a)=1/\max\{|\hat\theta_{t,i}-\hat\theta_{t,j}|,10^{-6}\}$.
\textsc{Arena2024} uses the sampling rule of \citet{Arena}, assigning weight
$w_t(a)=\sqrt{S_t(a)/n_t(a)}-\sqrt{S_t(a)/(n_t(a)+1)}$,
where $S_t(a)=\hat p_t(a)(1-\hat p_t(a))$ and $n_t(a)$ is the number of observations of the pair, floored at one.
\textsc{Arena2025} is a variant of the policy described by \citet{arena2025leaderboardillusion}: it assigns weight
$w_t(a)=v_t(i)v_t(j)$, where $v_t(i)=3$ for models in the top $30\%$ by current score and $v_t(i)=1$ otherwise.
Under every policy, the pair is chosen uniformly at random with probability $0.1$. Table \ref{tab: semiPolicy} confirms that the preceding conclusions hold across sampling policies: both IPW estimators maintain coverage near the nominal level, and PP-Rank reduces MSE relative to the labeled-only IPW baseline. The judge-only estimator loses coverage throughout, while the other methods without IPW retain coverage only under \textsc{Arena2024} because this policy allocates close to uniformly. %

\begin{table}[!ht]
\centering
\small
\begin{tabular}{lcccccc}
\hline
Policy &
\shortstack{Labeled-only\\(no IPW)} &
\shortstack{Labeled-only\\(IPW)} &
Judge-only & PPI++ & PP-Rank &
\shortstack{MSE\\reduction ($\uparrow$)} \\
\hline
\textsc{Greedy}      & 0.40 & 0.88 & 0.00 & 0.00 & 0.88 & 36\% \\
\textsc{Uncertainty} & 0.50 & 0.87 & 0.00 & 0.00 & 0.88 & 33\% \\
\textsc{Arena2024}   & 0.88 & 0.88 & 0.00 & 0.86 & 0.87 & 30\% \\
\textsc{Arena2025}   & 0.72 & 0.88 & 0.00 & 0.17 & 0.88 & 34\% \\
\hline
\end{tabular}
\caption{Results under four sampling policies. The first five numerical columns report minimum coverage of nominal 90\% confidence intervals across all 45 pairwise Bradley--Terry score contrasts, for which the target is 0.90. The two IPW estimators stay near nominal coverage, while methods designed for i.i.d.\ data undercover under nonuniform sampling. The final column reports PP-Rank's MSE reduction relative to the labeled-only IPW baseline. $T=10{,}000$, $\epsilon=0.1$, 1,000 trials. PP-Rank reduces MSE under every policy.}
\label{tab: semiPolicy}
\end{table}

\section{Synthetic Simulation}\label{sec: synthetic}

We conduct a fully synthetic simulation with $K=4$ models to be ranked.
Contexts $\bm X_t \sim N(\bm 0, I_{10})$ are 10-dimensional, and actions $A_t$ correspond to one of the ${4\choose 2} = 6$ pairs of models.
For $A_t = (j,k),$ the binary outcome $Y_t=1$ indicates a preference for item $j$ over item $k$, generated according to a contextual Bradley--Terry model
\(
\mathbb P(Y=1\mid A=(j,k),\bm X)=p_{(j,k)}(\bm X_{1:5}),
p_{(j,k)}(\bm x)=\sigma\big(\alpha_j-\alpha_k+\bm x^\top(\bm\gamma_j-\bm\gamma_k)\big),
\)
where $\bm X_{1:5}$ denotes the first five coordinates of $\bm X$, $\bm\alpha=(2,1,-1,-2)$, and $\bm\gamma_j$ is the $j$th row of 
\[
\Gamma=\begin{pmatrix}
 4.5 &  3.0 &  1.5 &  0   & -1.5\\
 0.6 & -1.2 &  1.2 &  0.6 &  0  \\
-0.4 &  0.2 & -0.2 &  0.4 &  0.2\\
-0.2 & -0.1 & -0.2 & -0.3 &  0
\end{pmatrix}.
\]
Only the first five coordinates of the context affect preferences; the remaining five, $\bm X_{6:10}$, are used below to define a poor predictor.
The decision-maker estimates the parameters $\beta_0^*,\beta_1^*,\beta_2^*,\beta_3^* \in \mathbb R$ of a misspecified working Bradley-Terry model
$$\mathbb P_{\bm\theta}(Y=1|A=(j,k)) = \sigma(\beta_j-\beta_k),$$
which assigns each item a single ability and omits the dependence on $\bm X.$
The target is defined under a uniform distribution over the six item pairs, with $\sum_{j=0}^3 \beta_j = 0$ imposed for identifiability.
We approximate the true value of $\bm\beta^*$ by Monte Carlo simulation, obtaining $\bm\beta^* = (0.54, 0.63, -0.25, -0.93).$

The decision-maker has access to AI models that predict preference probabilities, indexed by a quality parameter $\tau\in[0,1]$ interpolating between a perfect predictor and a poor one,
\[
f^\tau_{(j,k)}(\bm X)=(1-\tau)\,p_{(j,k)}(\bm X_{1:5})+\tau f^{\text{poor}}_{(j,k)}(\bm X).
\]
At $\tau=0$, the judge is perfect in the sense that it outputs the true conditional preference probability. The poor predictor applies the true rule, with the two models swapped, to the five coordinates that do not affect preferences:
\[
f^{\text{poor}}_{(j,k)}(\bm X)=p_{(k,j)}(\bm X_{6:10})=1-p_{(j,k)}(\bm X_{6:10}).
\]
It is independent of the true preference probability, so $\operatorname{corr}\big(f^\tau_{(j,k)}(\bm X),\,p_{(j,k)}(\bm X_{1:5})\big)=(1-\tau)/\sqrt{(1-\tau)^2+\tau^2}$ for every pair, and its average win rates are those of the reversed ranking. A judge-only estimate is therefore biased for every $\tau>0$.

At each time step, the decision-maker also receives $10$ independent unlabeled contexts $\tilde{\bm X}_i \sim N(\bm 0, I_{10}).$ The policy begins with a burn-in period that selects pairs uniformly at random until every pair has been compared 10 times.
Then, it selects model pairs using an $\epsilon$-greedy policy with $\epsilon = 0.1$.
With probability $\epsilon,$ it selects a pair uniformly at random, and with probability $1-\epsilon,$ it selects the pair consisting of the two models with the largest estimated strength according to the working model and labeled-only estimator, where item $j$ has estimated strength $\hat\beta_{j,t}$.

\subsection{Results}

\begin{figure}[h!]
    \centering
    \includegraphics[width=1\linewidth]{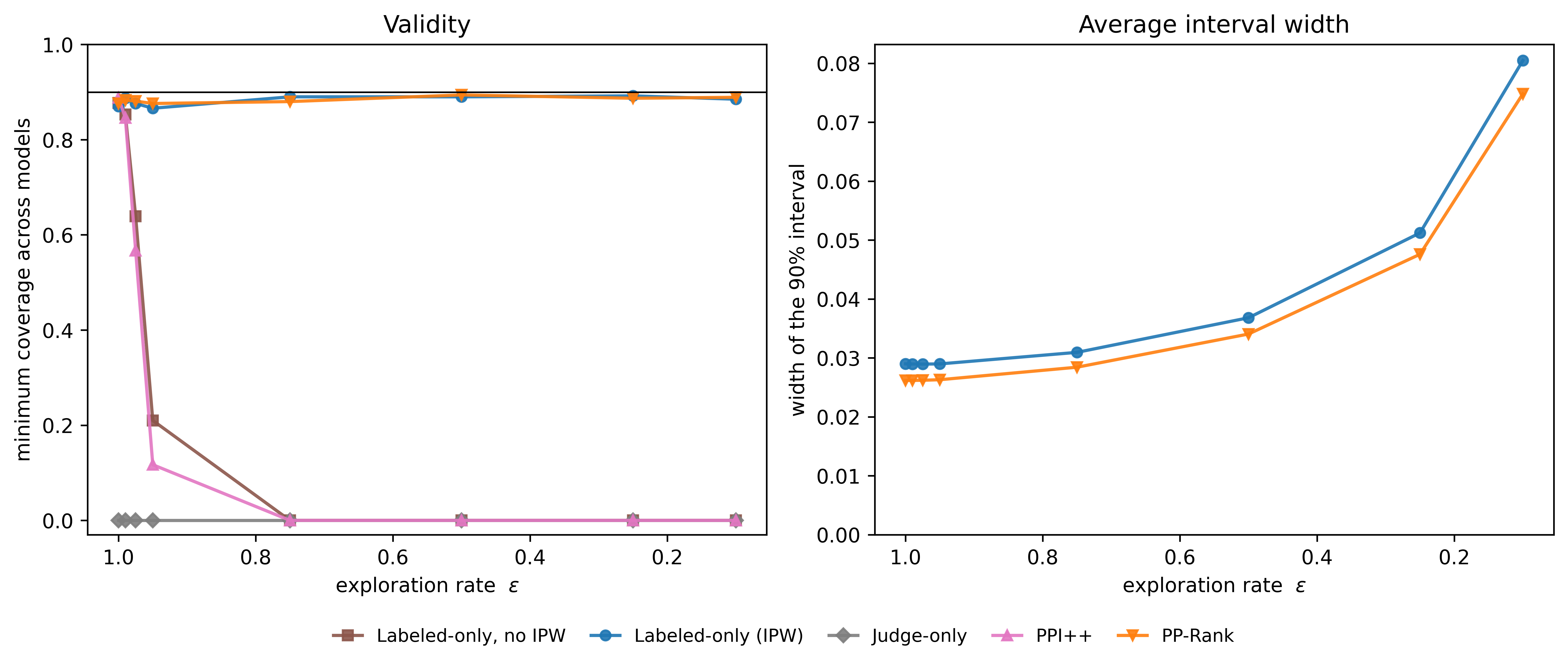}
    \caption{Left: minimum coverage across the four models' win rates at nominal 90\%. Right: average width of the same intervals for the two estimators that remain valid. Only the inverse-propensity-weighted estimators remain valid as exploration decreases, and PP-Rank's intervals are narrower at every exploration rate. We set $T=5000$ and $\tau=0.5$.}
    \label{fig: validitySynthetic}
\end{figure}

Figure~\ref{fig: validitySynthetic} shows that, for $\tau=0.5$,
PP-Rank and the labeled-only IPW baseline maintain coverage near nominal
as sampling becomes more adaptive, and PP-Rank produces narrower intervals.
Appendix~S5.2 presents the same plots
for perfect ($\tau=0$) and uninformative ($\tau=1$) judges,
along with the sampling-policy ablation, showing similar results.

We also evaluate the cost-reduction procedure developed in Section \ref{sec: costreduc}.
We take prices from four models on the Chatbot Arena cost-quality Pareto frontier as of September 15th, 2026: \$40, \$3, \$0.16 and \$0.07 per million tokens.
Item $j$ is assigned the $j$th price, and selecting pair $(i,j)$ incurs the sum of the two, so the low-cost rule always selects the pair of cheapest models. We repeat the experiments, but with PP-Rank we adjust the minimum sampling probability via
$\pi_{\min,t} = \max\left\{\widehat\pi^{PP}_{\min,t-1},\ \underline{\pi}\right\},
\underline{\pi} = 0.01,$
where $\widehat\pi^{PP}_{\min,t}$ solves \textup{(S23)} with a scalar $\lambda$. 

Table \ref{tab:cost-precision}
shows that PP-Rank substantially reduces evaluation costs,
particularly with a strong LLM judge, while maintaining
inferential precision comparable to the labeled-only IPW estimator in this setting,
as measured by MSE, CI width, and rank-set width.

\begin{table}[!ht]
\centering
\begin{tabular}{lcccc}
\hline
Method & \shortstack{Cost\\reduction ($\uparrow$)} & \shortstack{MSE\\ratio} & \shortstack{CI-width\\ratio} & \shortstack{Rank-set\\width ratio} \\
\hline
Labeled-only & --- & 1.00 & 1.00 & 1.00 \\
PP-Rank, $\tau=0$ & 52.6\% & 0.95 & 0.96 & 0.98 \\
PP-Rank, $\tau=0.25$ & 40.7\% & 0.98 & 0.98 & 0.99 \\
PP-Rank, $\tau=0.5$ & 14.3\% & 1.00 & 1.00 & 1.00 \\
PP-Rank, $\tau=0.75$ & 2.0\% & 1.00 & 1.00 & 1.00 \\
PP-Rank, $\tau=1$ & 0.2\% & 1.00 & 1.00 & 1.00 \\
\hline
\end{tabular}
\caption{PP-Rank reduces evaluation costs while maintaining precision comparable to the labeled-only IPW baseline, as its MSE, CI-width, and rank-set-width ratios relative to the baseline are all near 1. Results are averaged over 1,000 trials with $T=5{,}000$ and $N=50{,}000$.}
\label{tab:cost-precision}
\end{table}

\section{Discussion}

In this article, we introduce PP-Rank, a framework for valid and efficient LLM ranking that jointly leverages LLM-judge feedback, unlabeled prompts, and adaptively collected human preferences.
PP-Rank remains valid when evaluation platforms adaptively select model comparisons based on previous human feedback, and its data-driven weighting procedure exploits accurate LLM-judge predictions while limiting reliance on inaccurate ones.
We also show how LLM-judge feedback can relax the requirements on human sampling,
allowing evaluation platforms to prioritize lower-cost comparisons
without sacrificing inferential precision.

Semi-synthetic and synthetic experiments illustrate these benefits:
PP-Rank maintains valid inference under adaptive sampling, improves
precision and certification of ranking statements, and substantially
reduces evaluation costs. PP-Rank thus enables evaluation platforms
to make more reliable ranking statements from limited human feedback
while retaining flexibility in how model comparisons are selected.

Future work could study more general settings, such as dynamic and heterogeneous evaluations of LLMs. Examples include nonstationary leaderboards, where human preferences shift over time; LLM judges trained online using accumulated human feedback; and
personalized rankings for specific users or tasks.

\vspace{-0.2in}
\section*{Data Availability Statement}\vspace{-0.2in}
The UltraFeedback-Extended dataset  used in  data analysis is publicly available at \url{https://huggingface.co/datasets/ltg/ultrafeedback-extended}.

\vspace{-0.2in}
\section*{Supplementary Material}\vspace{-0.15in}
The supplement contains proofs %
and additional numerical experiments. %

\vspace{-0.3in}

\section*{Declaration of GenAI Use}\vspace{-0.15in}
The authors used GPT-6.1 Sol and Claude Opus 5.5 to assist with coding, proof verification, and manuscript editing.  The authors verified the accuracy and originality of the work and take full responsibility for the results.
\vspace{-0.15in}
\bibliographystyle{chicago}
\bibliography{references}

\end{document}